\documentclass{article}
\pdfoutput=1
\usepackage[nonatbib, preprint]{neurips_2021}

\usepackage[utf8]{inputenc} 
\usepackage[T1]{fontenc}    
\usepackage{hyperref}      
\usepackage{url}            
\usepackage{booktabs}       
\usepackage{amsfonts}       
\usepackage{nicefrac}       
\usepackage{microtype}     
\usepackage{xcolor}         
\usepackage{multirow}
\usepackage{bm}
\usepackage{amssymb}
\usepackage{amsfonts}
\usepackage{amsmath}
\usepackage{graphics}
\usepackage{graphicx}
\usepackage{amssymb}

\usepackage{caption}
\usepackage{subcaption}
\usepackage{wrapfig}

\title{Maximum Likelihood Estimation for Multimodal Learning with Missing Modality}

\author{%
  Fei~Ma  \\
  Tsinghua-Berkeley Shenzhen Institute\\
  Tsinghua University\\
  \texttt{mf17@mails.tsinghua.edu.cn} \\
   \And
 Xiangxiang~Xu  \\
 Department of EECS \\
  Massachusetts Institute of Technology\\
  \texttt{xuxx@mit.edu} \\
   \And
  Shao-Lun~Huang  \\
  Tsinghua-Berkeley Shenzhen Institute\\
  Tsinghua University\\
  \texttt{shaolun.huang@sz.tsinghua.edu.cn} \\
   \And
  Lin~Zhang  \\
  Tsinghua-Berkeley Shenzhen Institute\\
  Tsinghua University\\
  \texttt{linzhang@tsinghua.edu.cn} \\
}

\begin{document}

\maketitle

\begin{abstract}
Multimodal learning has achieved great successes in many scenarios. Compared with unimodal learning, it can effectively combine the information from different modalities to improve the performance of learning tasks.
In reality, the multimodal data may have missing modalities due to various reasons, such as sensor failure and data transmission error.
In previous works, the information of the modality-missing data has not been well exploited.
To address this problem, 
we propose an efficient approach based on maximum likelihood estimation to incorporate the knowledge in the modality-missing data.
Specifically, 
we design a likelihood function to characterize the conditional distribution of the modality-complete data and the modality-missing data, which is theoretically optimal.
Moreover, we develop a generalized form of the softmax function to effectively implement maximum likelihood estimation in an end-to-end manner. Such training strategy guarantees the computability of our algorithm capably.
Finally, we conduct a series of experiments on real-world multimodal datasets. 
Our results demonstrate the effectiveness of the proposed approach, even when 95\% of the training data has missing modality.

\end{abstract}

\section{Introduction}
Multimodal learning is an important research area, which builds models to process and relate information between different modalities \cite{baltruvsaitis2018multimodal}.
Compared with unimodal learning, multimodal learning can effectively utilize the multimodal data to achieve better performance.
It has been successfully used in many applications, such as multimodal emotion recognition \cite{soleymani2011multimodal}, multimedia event detection \cite{gan2015devnet}, and visual question-answering \cite{antol2015vqa}. 
With the emergence of big data, multimodal learning becomes more and more important to combine the multimodal data from different sources.

\begin{figure*}[]
	\centering
	\includegraphics[width=\linewidth]{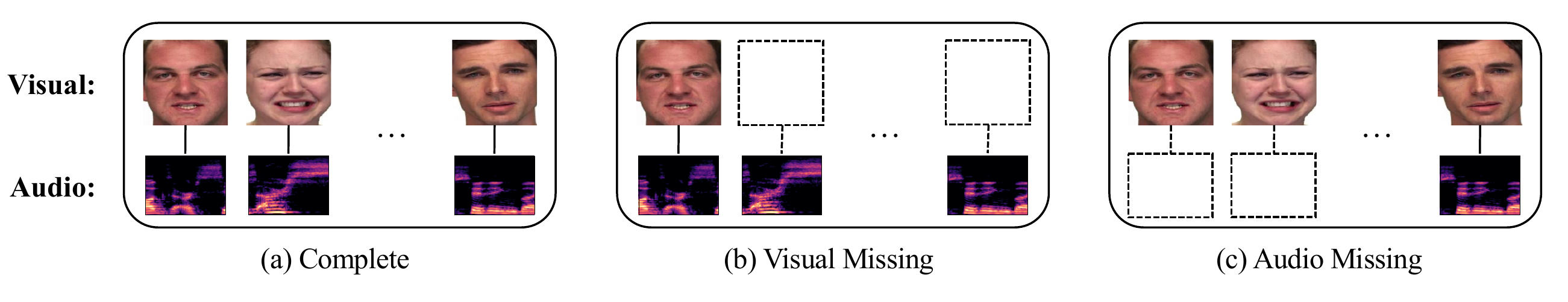}
	\caption{
	Examples of the multimodal data: (a) complete observations, (b) observations which may have missing visual modality, and (c) observations which may have missing audio modality.
	}
	\label{fig:missing_example}
\end{figure*}

A number of previous works \cite{tzirakis2017end,zhang2017learning,elliott2017findings,kim2020hypergraph,8999528} have achieved great successes based on complete observations
during the training process. However, 
in practice, the multimodal data may have missing modalities \cite{du2018semi,ma2021,ma2021smil}.
This may be caused by various reasons. For instance, 
the sensor that collects the multimodal data is damaged or the network transmission fails. Examples of the multimodal data are shown in Figure \ref{fig:missing_example}.

In the past years, researchers have proposed a few approaches to deal with modality missing.
A simple and typical way \cite{hastie2009elements} is
to directly discard the data with missing modalities. 
Since the information contained in the modality-missing data is neglected, such method often has limited performance.
Moreover, there are also approaches proposed to heuristically combine the information of the modality-missing data \cite{ma2021smil,8100011,chen2020hgmf,liu2021incomplete}. However, most of these works lack theoretical explanations, and these empirical methods are often implemented using
multiple training stages rather than an end-to-end manner, 
which lead to the information of the modality-missing data not being well exploited.

To tackle this problem, we propose an efficient approach based on maximum likelihood estimation to effectively utilize the modality-missing data.
To be specific, we present a likelihood function to characterize the conditional distribution of the modality-complete data and the modality-missing data, which is theoretically optimal.
Furthermore, we adopt a generalized form of the softmax function to efficiently implement our maximum likelihood estimation algorithm. Such training strategy guarantees the computability of our framework in an end-to-end scheme.
In this way, our approach can effectively leverage the information of the modality-missing data during the learning process, 
which has higher efficiency than previous works.
Finally, we perform several experiments on real-world multimodal datasets, including eNTERFACE’05 \cite{1623803} and RAVDESS \cite{livingstone2018ryerson}.
The results show the effectiveness of our approach in handling problems of modality missing.
To summarize, our contribution is three-fold:
\begin{itemize}
\item We design a likelihood function to learn the conditional distribution of the modality-complete data and the modality-missing data, which is theoretically optimal. 
\item We develop a generalized form of the softmax function to implement our maximum likelihood estimation framework in an end-to-end manner,
which is more effective than previous works.
\item We conduct a series of experiments on real-world multimodal datasets. The results validate the effectiveness of our approach, even when 95\% of the training data has missing modality.
\end{itemize}

\section{Methodology}
Our goal is to handle modality missing based on maximum likelihood estimation for effective multimodal learning. 
In the following, we first introduce the problem formulation, and then describe the details of our framework.

\subsection{Problem~Formulation}
\label{formulation}

In this paper, without loss of generality, we consider that the multimodal data has two modalities. 
Here, the random variables corresponding to these two modalities and their category labels are denoted as $X$, $Y$, and~$Z$, respectively.
In the training process, we assume that there are two independently observed datasets: modality-complete and modality-missing. 
We use $D_{XYZ} = \big\{ (x^{(i)}_{\rm{c}}, y^{(i)}_{\rm{c}}, z^{(i)}_{\rm{c}}) \big\}_{i=1}^{n_{\rm{c}}} $ to represent the modality-complete dataset, where $x^{(i)}_{\rm{c}}$ and $y^{(i)}_{\rm{c}}$ represent the two modalities of the \emph{i}-th sample of $D_{XYZ}$,
$z^{(i)}_{\rm{c}}$ is their corresponding category label, and the size of $D_{XYZ}$ is $n_{\rm{c}}$.
We then use $D_{XZ} = \big\{ (x^{(i)}_{\rm{m}}, z^{(i)}_{\rm{m}}) \big\}_{i=1}^{n_{\rm{m}}}$ to represent the modality-missing dataset, where the size of $D_{XZ}$ is $n_{\rm{m}}$.
In addition, we adopt $[D_{XYZ}]_{XY}$ to represent $\big\{ (x^{(i)}_{\rm{c}}, y^{(i)}_{\rm{c}}) \big\}_{i=1}^{n_{\rm{c}}}$. 
$[D_{XYZ}]_{Z}$, $[D_{XZ}]_{X}$, and $[D_{XZ}]_{Z}$ are expressed in the same way.
The multimodal data of $D_{XYZ}$ and $D_{XZ}$ are assumed to be i.i.d. generated from
the underlying distribution $Q_{XYZ}$.
By utilizing the knowledge of the 
modality-complete data and the modality-missing data, we hope our framework can predict the category labels correctly.

\begin{figure*}[t]
	\centering
	\includegraphics[width=\linewidth]{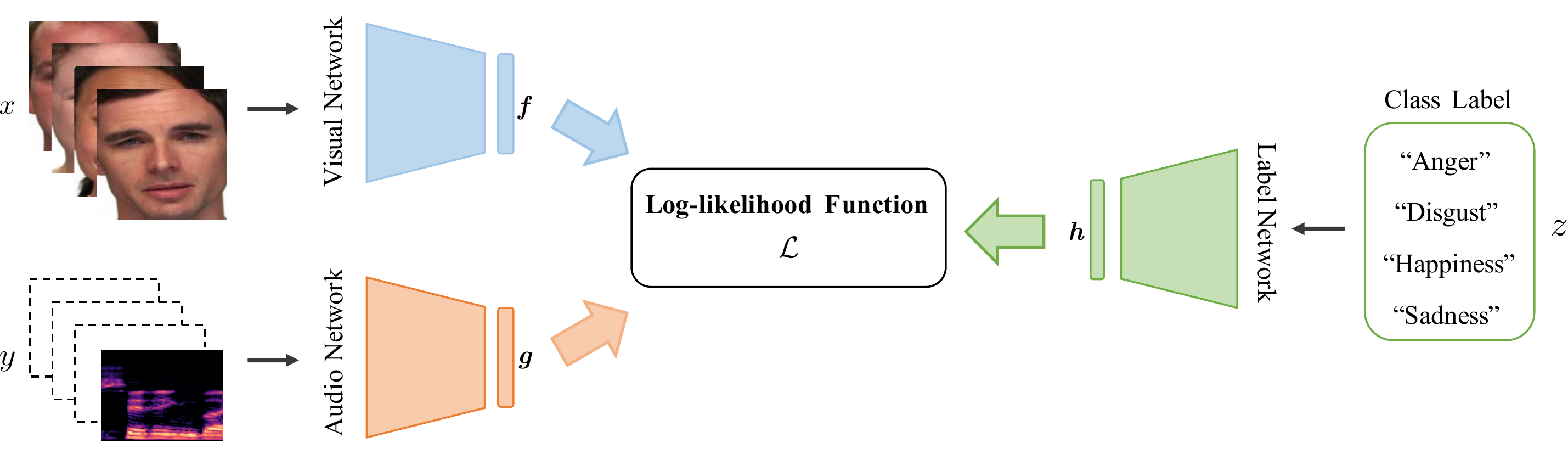}
	\caption{
	Our proposed system for multimodal learning with missing modality. In the training process, 
	we propose a log-likelihood function
	$\mathcal{L}$, as shown in Equation (\ref{conditional}), to learn the  conditional distribution of the modality-complete data and the modality-missing data.
By developing a generalized form of the softmax function, we implement our maximum likelihood estimation algorithm in an end-to-end manner, which has high efficiency.
	}
	\label{fig:mle_missint_modalities}
\end{figure*}

\subsection{Maximum Likelihood Estimation for Missing Modality}
\label{model}

In this section, we first present how to design a likelihood function to learn the conditional distribution of the modality-complete data and the modality-missing data.
Then, we show that by adopting a generalized form of the softmax function, we design a training strategy to effectively implement our algorithm.
\subsubsection{Likelihood Function Analyses}

Maximum likelihood estimation is a statistical method of 
using the observed data to estimate the underlying distribution by maximizing the likelihood function.
The estimated distribution makes the observed data most likely \cite{myung2003tutorial}. With this idea, we study the likelihood function on datasets $D_{XYZ}$ and $D_{XZ}$.
For the classification task, the conditional likelihood is commonly used. 
Inspired by this,
we analyze the conditional likelihood, which can be represented as:

\begin{equation} 
\label{likelihood function}
\begin{aligned}
    \ell  &  \triangleq  \mathbb{P} \left( [D_{XYZ}]_{Z}, [D_{XZ}]_{Z} \mid [D_{XYZ}]_{XY}, [D_{XZ}]_{X} ; Q_{XYZ} \right)\\
    & \overset{\textbf{a}}{=}  \mathbb{P} \left( [D_{XYZ}]_{Z} \mid [D_{XYZ}]_{XY} ; Q_{XYZ} \right) \cdot \mathbb{P} \left( [D_{XZ}]_{Z} \mid [D_{XZ}]_{X} ; Q_{XYZ} \right) \\
    & \overset{\textbf{b}}{=}  \prod_{(x, y, z)\in D_{XYZ}}Q_{Z|XY}(z|xy) \cdot \prod_{(x, z)\in D_{XZ}}Q_{Z|X}(z|x) 
\end{aligned}
\end{equation}

where the step $\textbf{a}$ follows from the fact that datasets $D_{XYZ}$ and $D_{XZ}$ are observed independently, and the step $\textbf{b}$ is due to that samples in each dataset are i.i.d.
$Q_{Z|XY}$ and $Q_{Z|X}$ are conditional distributions of $Q_{XYZ}$.
In this way, we show the likelihood function using the information of $D_{XYZ}$ and $D_{XZ}$.
Then, we use the negative log-likelihood as the loss function to train our deep learning network, i.e.,
\begin{equation}
\label{conditional}
\mathcal{L} \triangleq -\log \ell = - \sum_{(x, y, z)\in D_{XYZ}}\log Q_{Z|XY}(z|xy) - \sum_{(x, z) \in D_{XZ} }\log Q_{Z|X}(z|x)
\end{equation}
It is worth noting that in \cite{daniels1961asymptotic}, maximum likelihood estimation is proved to be an asymptotically-efficient strategy. Therefore, 
the theoretical optimality of our method is guaranteed to deal with modality missing. 

To optimize 
$\mathcal{L}$, we use deep neural networks to extract the $k$-dimensional feature representations from the observation $(x,y,z)$, which are represented as 
$\bm{f}(x)=\left [ f_1(x), f_2(x),\cdots , f_k(x) \right ]^{\textup{T}}$, $\bm{g}(y)=\left [ g_1(y), g_2(y),\cdots , g_k(y) \right ]^{\textup{T}}$, and~$\bm{h}(z)=\left [ h_1(z), h_2(z),\cdots , h_k(z) \right ]^{\textup{T}}$, respectively.
We then utilize these features to learn $Q_{Z|XY}$ and $Q_{Z|X}$ in $\mathcal{L}$. Our framework is shown in Figure \ref{fig:mle_missint_modalities}. 

In this way, we show the log-likelihood function $\mathcal{L}$. By characterizing the conditional distribution of the modality-complete data and modality-missing data, it efficiently leverages the underlying structure information behind the multimodal data, which constitutes the theoretical basis of our framework.

\subsubsection{Maximum Likelihood Estimation Implementation}
\begin{figure*}[]
	\centering
	\includegraphics[width=\linewidth]{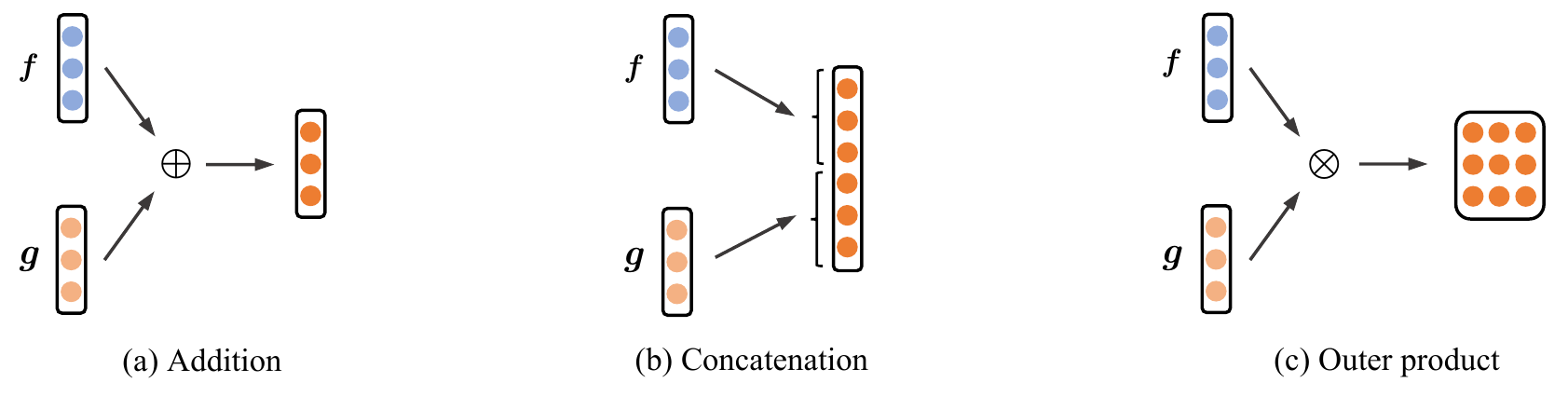}
	\caption{
Three forms of $\bm{\phi}$ are studied:
(a) addition, i.e., $\bm{\phi}(\bm{f},\bm{g})\triangleq \bm{f} + \bm{g}$, (b) concatenation, i.e., $\bm{\phi}(\bm{f},\bm{g})\triangleq [\bm{f}^{\text{T}}, \bm{g}^{\text{T}}]^{\text{T}} $, and (c) outer product, i.e., $\bm{\phi}(\bm{f},\bm{g}) \triangleq \textbf{vec}(\bm{f} \otimes \bm{g})  $, where $\textbf{vec}$ represents the vectorization of outer product. 
	}
	\label{fig:example_missing_modalities}
\vspace{-12pt}
\end{figure*}

However, the log-likelihood function $\mathcal{L}$ in Equation (\ref{conditional}) cannot be used directly, which is mainly due to two facts. Firstly, the representations of the high-dimensional data and how to model them are complicated.
Secondly, since $Q_{Z|XY}$ and $Q_{Z|X}$ in $\mathcal{L}$ are related, how to design models to learn their relationships is difficult.
To address these two issues, we develop a generalized form of the softmax function to describe $Q_{XYZ}$ as follows:
\begin{equation}
\label{q_xyz}
    Q_{XYZ}(x, y,z) = \frac{R_X(x)R_Y(y)R_Z(z)\exp(\bm{\phi}^{\text{T}} (\bm{f}(x),\bm{g}(y))\bm{h}(z))}{\sum_{x', y', z'}R_X(x')R_Y(y')R_Z(z')\exp( \bm{\phi}^{\text{T}} (\bm{f}(x'),\bm{g}(y'))\bm{h}(z'))}
\end{equation}

where $R_X$, $R_Y$, and $R_Z$ represent the empirical distributions obtained by using all observed samples of the variables $X$, $Y$, and $Z$, respectively.
$\bm{\phi}(\bm{f},\bm{g})$ represents the function to fuse features $\bm{f}$ and $\bm{g}$.
We study three forms of $\bm{\phi}$ to investigate its effect in our framework, as shown in Figure \ref{fig:example_missing_modalities}.

In this way, we show the underlying distribution $Q_{XYZ}$ by adopting a generalized form of the softmax function, which has the following two benefits. Firstly, by depicting the representation of $Q_{XYZ}$, we can further deduce $Q_{Z|XY}$ and $Q_{Z|X}$ directly. It guarantees our algorithm can be effectively implemented in an end-to-end manner. Secondly, it avoids giving the expressions of $Q_{Z|XY}$ and $Q_{Z|X}$ and modeling the relationship between them.
In fact, it is hard to compute these marginalized distributions since the correlation between the high-dimensional data can be rather complex. 
In addition, it has been shown in \cite{8613303} for the case with two random variables, the generalized version of softmax we adopt is equivalent to the standard softmax function.

The conditional distributions $Q_{Z|XY}$ and $Q_{Z|X}$ can be easily obtained correspondingly as follows:
\begin{equation}
\label{q_z|xy}
Q_{Z|XY}(z|xy) = R_Z(z)\frac{\exp(\bm{\phi}^{\text{T}}(\bm{f}(x),\bm{g}(y)) \bm{h}(z))}{\sum_{z'}R_Z(z')\exp(\bm{\phi}^{\text{T}}(\bm{f}(x),\bm{g}(y)) \bm{h}(z'))}
\end{equation}
and
\begin{equation}
\label{q_z|x}
Q_{Z|X}(z|x) = R_Z(z)\frac{\sum_{y'}R_Y(y')\exp(\bm{\phi}^{\text{T}}(\bm{f}(x),\bm{g}(y')) \bm{h}(z))}{\sum_{z'}R_Z(z')\sum_{y'}R_Y(y')\exp(\bm{\phi}^{\text{T}}(\bm{f}(x),\bm{g}(y')) \bm{h}(z'))}
\end{equation}
It is worth pointing out that when we compute $Q_{Z|X}$ in Equation (\ref{q_z|x}), we need to use the information of the modality $y$. Since in the training process, the modality $y$ of the dataset $D_{XZ}$ is missing, 
we query all possible values of modality $y$ on $D_{XYZ}$ to compute $Q_{Z|X}$.
This can be regarded as using the modality-complete dataset to complement the modality-missing dataset.

We then plug Equation (\ref{q_z|xy}) and Equation (\ref{q_z|x}) into Equation (\ref{conditional}). In this way, we can use neural networks to learn features $\bm{f}$, $\bm{g}$, and $\bm{h}$ from $D_{XYZ}$ and $D_{XZ}$ for the classification task.
It does not need to complement the data before performing the classification task. 
Additionally, our objective function is a unified structure. Unlike previous works \cite{ma2021,ma2021smil}, it does not bring hyperparameters which need to be manually adjusted.
These factors guarantee the implementation of our approach is more efficient than previous methods.

\section{Experiments}
In this section, we first describe the real-world multimodal datasets used in our experiment, then explain the data preprocessing and baseline methods, and finally give the experimental results to show the effectiveness of our approach.

\subsection{Datasets}
\label{datasets}
We perform experiments on two public real-world multimodal datasets: eNTERFACE’05 \cite{1623803} and RAVDESS \cite{livingstone2018ryerson}. 
eNTERFACE’05 is an audio-visual emotion database in English. It contains 42 subjects eliciting the six basic emotions: anger, disgust, fear, happiness, sadness, and surprise. 
There are 213 samples for happiness, and 216 samples for each of the remaining emotions. Each recorded sample is in the video form, where the frame rate is 25 frames per second and the audio sampling rate is 48 kHz. 

RAVDESS is a multimodal database of emotional speech and song, which consists of 24 professional actors in a neutral North American accent. Here, we use the speech part, which includes calm, happy, sad, angry, fearful, surprise, and disgust expressions. Each recording is also in the video form. The frame rate is 30 frames per second. The audio is sampled at 48 kHz. Similar to the eNTERFACE’05 dataset, we only consider six basic emotions, each of which has 192 samples.

\subsection{Data Preprocessing and Experimental Settings}
\label{preprocessing}
We perform data preprocessing on these two datasets.
We split each video sample into segments of the same length. Then, we extract visual and audio data from these segments.
For each video sample, we obtain 30 segments, each of which has a duration of 0.5 seconds.
We then take the central frame as the visual data from each segment. In addition, we extract the log Mel-spectrogram of each segment as the audio data. The spectrum representation is similar to the RGB image.
We feed these segmented data into our model to obtain the classification result of segment level. Then we average the results of all segments belonging to the same video to predict the category label of the video level.

On each dataset, we split all data into three parts: training set, validation set, and test set. Their proportions are 70\%, 15\%, and 15\%.
The cases of incomplete audio modality and incomplete visual modality are separately studied.
In these two scenarios, we investigate the following missing rates: 50\%, 80\%, 90\%, and 95\%.
It is worth noting that to the best of our knowledge, we are the first to consider these settings with high missing rates.
We conduct experiments on different conditions  
to verify that our approach has a good generalization capability to deal with incomplete modalities. 
On the contrary, previous works only assume a certain modality is incomplete.
Following \cite{du2018semi,yu2020optimal,du2021multimodal}, we assume that in the inference phase, the test data is modality-complete.
Therefore, we can directly use Equation (\ref{q_z|xy}) to predict the class label of the given test data.
Finally, we run each experiment five times and report the average test accuracy to evaluate the performance of our method and some baseline methods.
All experiments are implemented by Pytorch \cite{paszke2019pytorch} on a NVIDIA TITAN V GPU card.

\subsection{Baseline Methods}
To show the effectiveness of our method, we compare our approach with the following methods which can also handle missing modalities to some extent.
\begin{itemize}
\item Discarding Modality-incomplete Data (Lower Bound): 
One of the simplest strategies to handle modality missing is to directly discard the modality-incomplete data, and then only use the modality-complete data for the classification task. 
This method does not use the information of the data with missing modalities. In our maximum likelihood estimation model, 
this is equivalent to calculating $Q_{Z|XY}$ without calculating $Q_{Z|X}$. 
Therefore, this method can also be used as the ablation study of our method. 

\item Hirschfeld-Gebelein-Renyi (HGR) Maximal Correlation \cite{hirschfeld1935connection,gebelein1941statistische,renyi1959measures}: HGR maximal correlation is a multimodal learning method based on the statistical dependence between different modalities. 
It has been successfully used for semi-supervised learning \cite{ma2021,app10207239,wang2019efficient}.
Here, we use it further to deal with incomplete modalities. 
For the data on $D_{XYZ}$, we extract the maximal correlation between $x$, $y$, and $z$. For the data on $D_{XZ}$, we extract the maximal correlation between $x$ and $z$. 
\item Zero Padding (ZP): 
Padding the feature representations of the missing modality with zero is a widely used way to copy with incomplete modalities \cite{jo2019cross,chen2020multi,shen2020memor}. 
For this method, we consider two forms of $\bm{\phi}$ to fuse features $\bm{f}$ and $\bm{g}$: addition and concatenation. The reason why the form of outer product is not studied here is that if the feature of one modality is zero, the outer product of it and the non-zero feature of another modality is also zero, which leads to the result that the modality-missing data is useless.
\item Autoencoder1 (AE1):
The autoencoder is an architecture to learn feature representations from training data in an unsupervised way. 
Some previous approaches try to use autoencoders to complement missing modalities \cite{8100011,liu2021incomplete,jaques2017multimodal,pereira2020reviewing}. 
Following these works, on the modality-complete dataset $D_{XYZ}$, we use modality $x$ as the input of the autoencoder to reconstruct modality $y$. Then we use the trained autoencoder to predict the modality $y$ on $D_{XZ}$ to impute the training data. Then we use the imputed data to perform the classification task. It is worth noting that the autoencoder used to deal with missing modality 
has several stages while our method is end-to-end.

\item Autoencoder2 (AE2):
In AE1, the data of $D_{XZ}$ are not involved in the training process of the autoencoder. Inspired by the self-training approach \cite{yarowsky1995unsupervised,mcclosky2006effective}, in each iteration, we predict the modality $y$ on $D_{XZ}$ as the pseudo value for the next iteration. 
In this way, the information of the modality-missing dataset can be integrated into the autoencoder to a certain extent. Here, we call this structure AE2.
\end{itemize}
\vspace{-8pt}

\subsection{Experimental Results}

In this section, we demonstrate the effectiveness of our method in two aspects. 
Firstly, we show that our method achieves high performance in tackling modality missing, even when the missing rate reaches 95\%. 
Secondly, we show that our method has higher efficiency than the autoencoder methods.

\begin{table}[t]
  \caption{The classification performance with missing modality
  on the eNTERFACE’05 dataset. }
  \label{audio-visual-missing-table}
  \centering
  \resizebox{\textwidth}{!}{
\begin{tabular}{cccccccc}
 \toprule
\multirow{2}{*}{Method} & \multicolumn{3}{c}{Visual Missing}                               &                      & \multicolumn{3}{c}{Audio Missing}                                 \\ \cmidrule(r){2-4} \cmidrule(r){6-8} 
                        & 80\%                 & 90\%                 & 95\%                 &                      & 80\%                 & 90\%                 & 95\%                 \\ 
                       \midrule
                        Lower Bound \cite{hastie2009elements} (Addition) 
                        &   46.91                   &    35.26                  &26.39       & 
                        &     50.93           &    35.26                  &  27.53                    \\
                        Lower Bound \cite{hastie2009elements} (Concatenation)
                        & 46.49                      & 36.39                     & 27.11      & 
                        &    46.29            &     33.71                 &   27.84                   \\
                        Lower Bound \cite{hastie2009elements} (Outer product)
                        &  42.78                    & 37.53                     & 26.91      & 
                        &   48.14             &  34.95                    &  28.56                    \\
                        HGR Maximal Correlation \cite{ma2021,app10207239,wang2019efficient} (Addition)
                        &  58.97                    &   59.69                   &  41.34     & 
                        &   77.32             &  74.12                    &   54.95                   \\
                        HGR Maximal Correlation \cite{ma2021,app10207239,wang2019efficient} (Concatenation)
                        &   63.51                   &   57.84                   & 41.34      & 
                        &   79.18             &    75.67                  &   57.42                   \\
                        HGR Maximal Correlation \cite{ma2021,app10207239,wang2019efficient} (Outer product)
                        &  64.64                    &   59.69                   & 49.90      & 
                        &   77.94             &  76.29                    &  55.46                    \\
                        ZP \cite{jo2019cross,chen2020multi,shen2020memor} (Addition)
                        &   69.07                   &    67.84                 & 58.66      & 
                        &   80.41             & 78.35                     &  76.49                    \\
                        ZP \cite{jo2019cross,chen2020multi,shen2020memor} (Concatenation)
                        &  68.76                    &  67.11                    & 60.21      & 
                        &   80.93             &    78.25                  &  76.70                    \\
                        Ours (Addition)
                        &    \textbf{72.37}                &  \textbf{71.65}                   & \textbf{66.29}      & 
                        &  81.24              &   80.31                   &   79.38                   \\
                        Ours (Concatenation)
                        & 72.27                   & 70.82                     &  64.74     & 
                        & 81.24               &    80.21                  &   79.79                   \\
                        Ours (Outer product)
                        &  72.06                   &  71.13                    & 66.08      & 
                        &    \textbf{81.65}            &  \textbf{81.03 }                   &    \textbf{80.31}                  \\
\bottomrule
\end{tabular}}
\end{table}

\begin{wrapfigure}{r}{0.5\textwidth}
  \centering
    \includegraphics[width=0.48\textwidth]{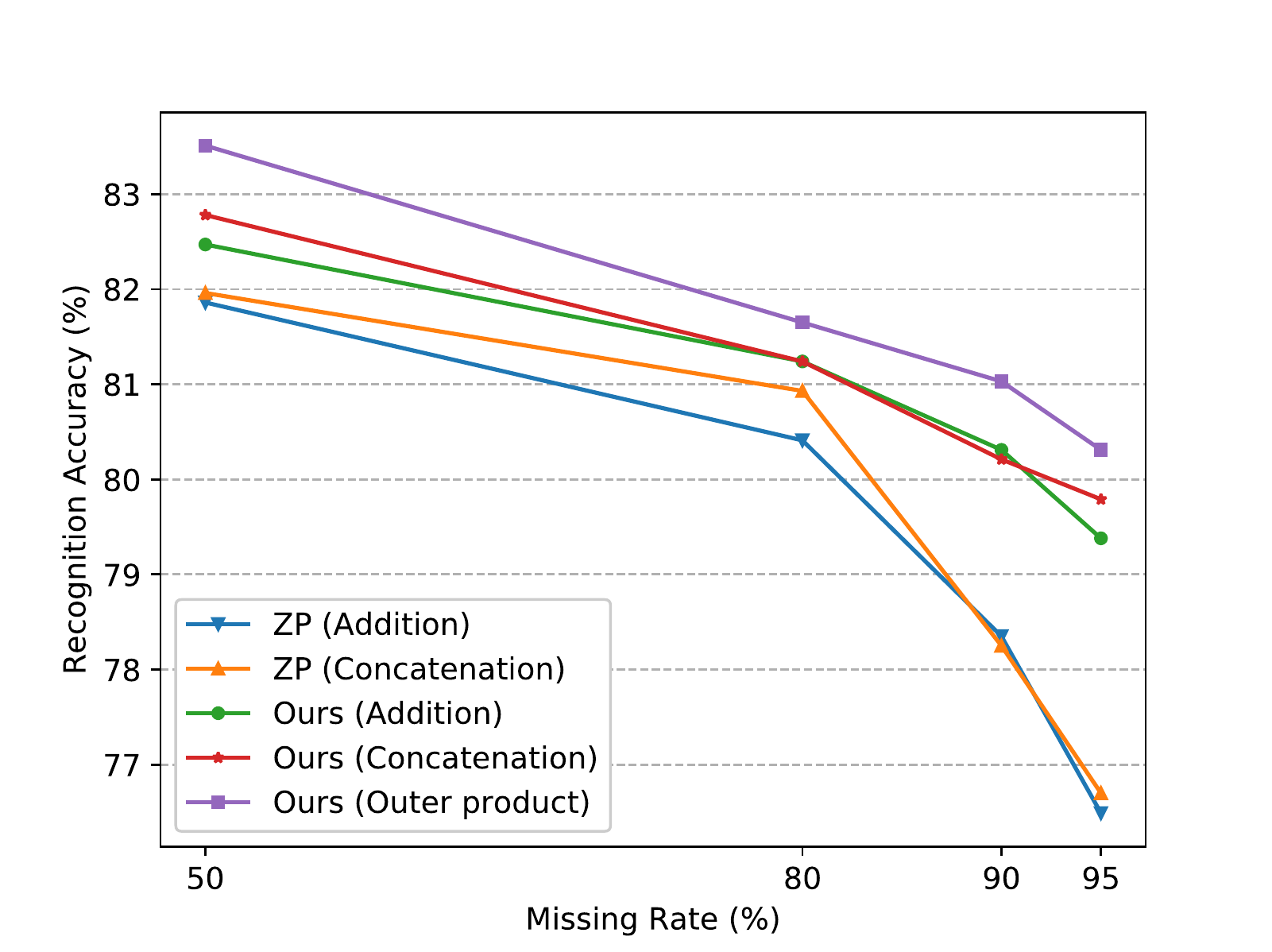}
  \caption{The tendency of ZP and ours as the missing rate increases when audio modality is missing on the eNTERFACE’05 dataset.}
  \label{zp_and_our}
\end{wrapfigure}

We first conduct emotion classification experiments on the eNTERFACE’05 dataset to compare our method with other end-to-end ones. 
We make the audio modality and the visual modality missing respectively. 
In each of these two scenarios, we set the missing rate to 80\%, 90\%, and 95\%.
The raw data and the corresponding labels are used as the input of our network. 
We adopt ResNet-50 \cite{7780459} as backbones to extract features from audio and visual modalities. 
In addition, we transform the label into the one-hot form and then get the corresponding label features using a fully connected layer. 
The whole network is trained together. 
For the fair comparison, different methods are set to have the same structure.
We report the classification accuracy of each method in each setting.
The results are shown in Table \ref{audio-visual-missing-table}.
In particular, when audio modality is missing, 
we analyze the tendency of ZP and ours as the missing rate increases, as shown in Figure \ref{zp_and_our}.

We have the following summarizations from Table \ref{audio-visual-missing-table} and Figure \ref{zp_and_our}: (1) The methods of HGR maximal correlation, ZP, and ours can improve the classification performance compared to the Lower Bound method which only uses the modality-complete data. 
Our method achieves the best performance, especially with $\bm{\phi}$ in the forms of outer product and addition. 
This shows that our method based on  maximum likelihood estimation can overcome modality missing effectively compared with other methods.
(2) When the visual modality is missing, the classification accuracy is lower than that when the audio modality is missing, indicating that the visual modality has a more significant contribution to the classification performance, which is consistent with the previous works \cite{zhang2017learning,app10207239}.
(3) When the missing rate increases, the classification accuracies of different methods decrease. Compared with other methods, the accuracy of our method decreases more slowly. This shows that our method is more capable of coping with missing modalities.
(4) For our approach, the ways to fuse features $\bm{f}$ and $\bm{g}$ with outer product and addition performs better than the way with concatenation. This indicates that in different scenarios, the discrimination ability of the learned feature representations is different. 
We need to design the appropriate form of $\bm{\phi}$ to fuse features in our framework.
(5) The method of HGR maximal correlation can deal with modality missing to a certain extent. 
However, it only focuses on the statistical dependence between different modalities and does not make full use of the information of different types of data, so its classification performance is worse than ZP and ours.

\begin{figure*}[t]
	\centering
	\includegraphics[width=\linewidth]{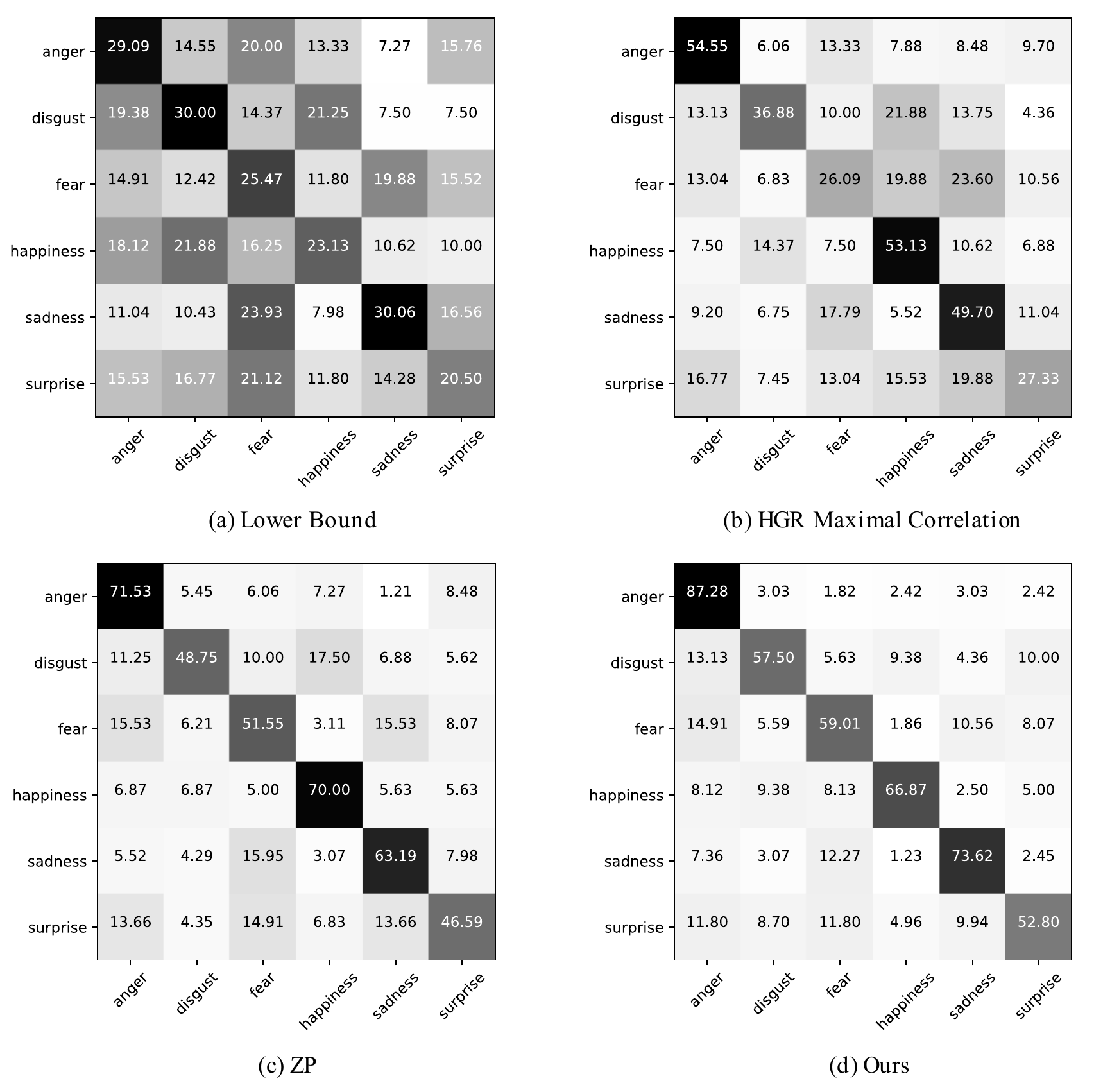}
	\caption{
	The confusion matrices of different methods on the eNTERFACE’05 dataset.
	}
	\label{fig:confusion}
\end{figure*}

In addition, we show the classification confusion matrices using the methods of Lower Bound, HGR maximal correlation, ZP, and ours
when the missing rate of visual modality reaches 95\% on the eNTERFACE’05 dataset, as shown in Figure \ref{fig:confusion}. 
It can be seen that the classification accuracy of each type of emotion using the Lower Bound method is not high since that it only combines the information from the modality-complete data. 
Compared with the Lower Bound method, HGR maximal correlation and ZP can improve the recognition accuracy of each type of emotion. 
The overall classification performance of ZP is lower than ours, but the classification accuracy of ``happiness'' is slightly higher than ours.
This shows that our method is more efficient to exploit the information from most emotions for the classification task.

\begin{wrapfigure}{r}{0.5\textwidth}
  \centering
    \includegraphics[width=0.5\textwidth]{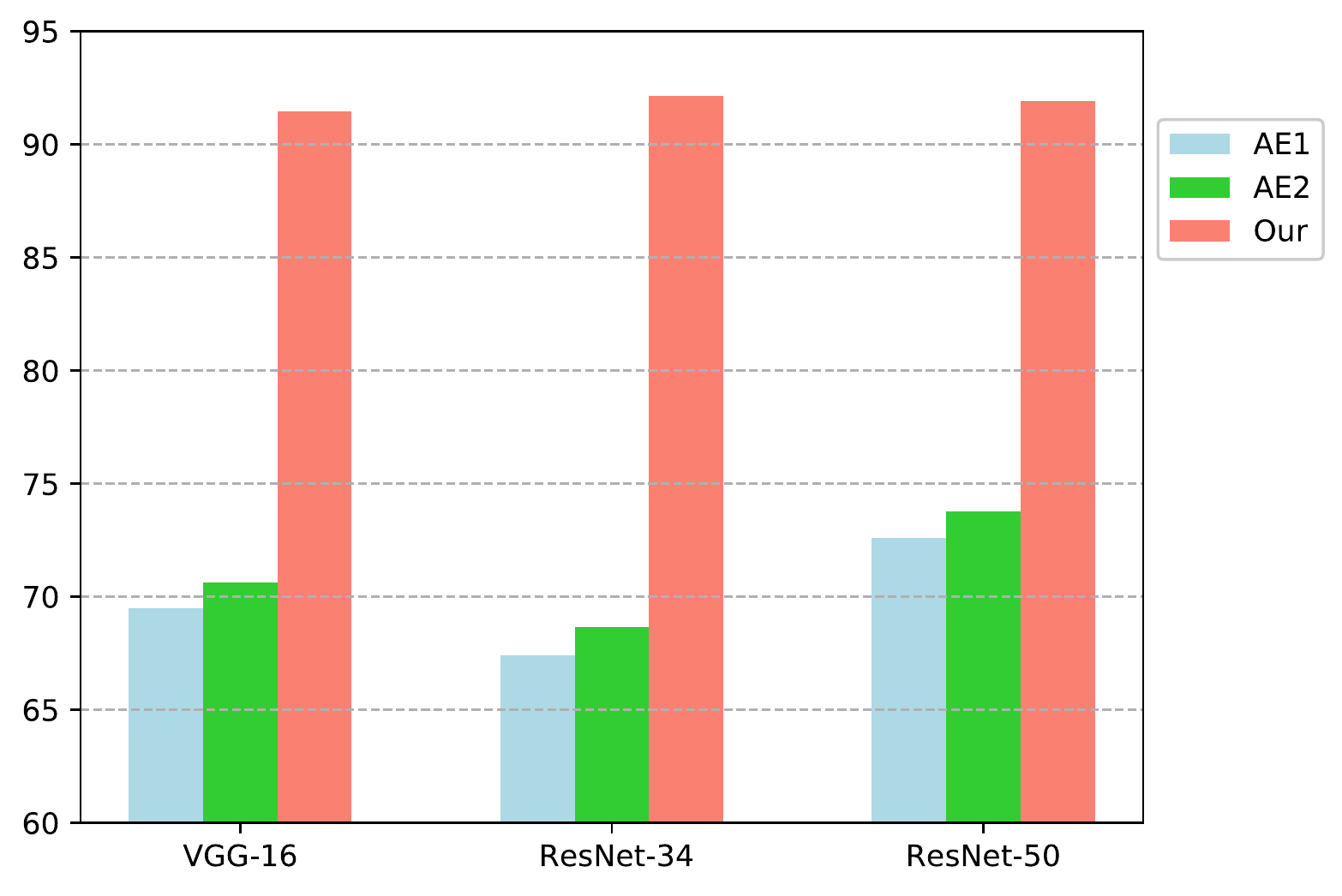}
  \caption{The performance comparison of different methods when 50\% of the training data has missing audio modality on the RAVDESS dataset.}
  \label{our_ae}
\end{wrapfigure}
We then compare our method with the method using autoencoders on the RAVDESS dataset to demonstrate that our method has high efficiency.
The method using autoencoders needs to be designed to reconstruct one modality using another modality.
It is difficult for autoencoders if we directly use the raw data as the input to perform this kind of cross-modal generation task. 
Therefore, we use some pre-trained networks, including VGG-16 \cite{simonyan2014very}, ResNet-34 and ResNet-50, to extract audio features and visual features from the raw data as the input of our model and the autoencoder. 
In other words, we reconstruct the features of different modalities here, and do not reconstruct the raw data. 
After reconstructing the features using the autoencoder method, we use the imputed feature for classification. We conducte experiments with AE1 and AE2 respectively. Correspondingly, we also use our method to classify the extracted features. 
We report the classification accuracy of each method within the same number of epochs to compare the efficiency of different methods.
The experimental results are shown in Table \ref{ravdess} and Figure \ref{our_ae}.

\begin{table}[t]
  \caption{The classification performance with missing modality
  on the RAVDESS dataset. }
  \label{ravdess}
  \centering
\begin{tabular}{ccccclccc}
 \toprule
\multicolumn{2}{c}{\multirow{2}{*}{Method}} & \multicolumn{3}{c}{Visual Missing} &  & \multicolumn{3}{c}{Audio Missing} \\ \cmidrule(r){3-5} \cmidrule(r){7-9} 
\multicolumn{2}{c}{}                        
& 50\%      & 80\%      & 90\%      &  & 50\%       & 80\%      & 90\%      \\ 
\midrule
\multirow{3}{*}{VGG-16}         
& AE1       
&  62.66         &  57.92         &    48.67       &  &  69.48          &   68.67        &   66.59        \\
& AE2       
&   66.24        &  58.15        &  48.90         &  &   70.64         &   69.48        &   66.59        \\
& Ours      
&   78.84        &  61.16         &  49.71         &  &  91.45          &    89.36       &  87.51         \\ 
\midrule
\multirow{3}{*}{Resnet-34}      
& AE1       
&   60.92        &   53.64        &   47.17        &  &   67.40         &  62.66         &   62.43        \\
& AE2       
&  62.31         &    54.10       &  49.60         &  &   68.67         &   63.47        &  62.77         \\
& Ours      
&  80.46         &    64.62       &  52.37         &  &  \textbf{92.14}          &  \textbf{90.06}         &   86.59        \\ 
\midrule
\multirow{3}{*}{Resnet-50}      
& AE1       
&   67.05        & 60.12          &   54.45        &  & 72.60          &  70.40         & 68.32           \\
& AE2       
&   69.13        &   61.16       &  50.64         &  &  73.76          & 70.06          & 66.94          \\
& Ours      
&   \textbf{84.05}       & \textbf{68.90}          &   \textbf{57.11}        &  & 91.91           & 89.48          & \textbf{88.79}          \\ 
\bottomrule
\end{tabular}
\end{table}

We have the following observations From Table \ref{ravdess} and Figure \ref{our_ae}:
(1) In each scenario, the classification accuracy of our method is higher than that of AE1 or AE2 within a certain number of epochs, which shows that our method has higher efficiency.
(2) In most cases, the classification accuracy of AE2 is generally higher than that of AE1, especially when the modality missing is not serious. 
This shows that if there are more modality-complete data for training, the autoencoder with self training can handle missing modalities more effectively for classification.
(3) When the size of modality-complete data increases, the classification accuracy of our method increases faster than that of AE1 and AE2. 
This may be owing to that our method is more efficient than AE1 and AE2 when combining the modality-complete data for classification.
(4) For our method, when the visual modality is missing, the classification accuracy using the features extracted by ResNet-50 is higher than that using VGG-16 and ResNet-34. When the audio modality is missing, in most settings, the classification accuracy using ResNet-34 is higher than that using VGG-16 and ResNet-50. 
This indicates that in different settings with missing modalities, we should adopt appropriate networks to extract features to make the features have high discrimination ability.

\section{Related Works}
Multimodal learning has achieved great successes in many applications.
An important topic in this field is multimodal representations \cite{baltruvsaitis2018multimodal,8970556}, which learn feature representations from the multimodal data by using the information of different modalities. 
How to learn good representations is investigated in \cite{ngiam2011multimodal,wu2014exploring,pan2016jointly,xu2015jointly}.
Another important topic is multimodal fusion \cite{atrey2010multimodal,poria2017review}, which combines the information from different modalities to make predictions.
Feature-based fusion is one of the most common types of multimodal fusion.
It concatenates the feature representations extracted from different modalities. 
This fusion approach is adopted by previous works \cite{tzirakis2017end,zhang2017learning,castellano2008emotion,zhang2016multimodal}.

To copy with the problem of modality missing for multimodal learning, a few methods have been proposed.
For example, 
in \cite{ma2021smil}, Ma et al. propose a Bayesian meta learning framework to perturb the latent feature space so that embeddings of single modality can approximate embeddings of full modality.
In \cite{8100011}, Tran et al. propose a cascaded residual autoencoder for imputation with missing modalities, which is composed of a set of stacked residual autoencoders that iteratively model the residuals.
In \cite{chen2020hgmf}, Chen et al. propose a heterogeneous graph-based multimodal fusion approach to enable multimodal fusion of incomplete data within a heterogeneous graph structure.
In \cite{liu2021incomplete}, Liu et al. propose an autoencoder framework to complement the missing data in the kernel space while taking into account the structural information of data and the inherent association between multiple views.

The above methods can combine the information of the modality-missing data to a certain extent. 
However, our method is more effective. 
The reason lies in the following two facts.
Firstly, by efficiently exploiting the likelihood function to learn the conditional distribution of the modality-complete data and the modality-missing data,
our method has a theoretical guarantee, which is skipped by previous works.
Secondly, the training process of our approach is more concise and flexible, while the training process of the above methods is relatively cumbersome.

In addition, 
it is worth noting that in \cite{tsai2018learning,9351737,9341029,pham2019found}, 
the multimodal data is assumed to be complete during the training process, and modality missing only occurs during the testing stage. 
These approaches make it hard to deal with missing modalities in the training phase, which may lead to limited performance.

\section{Conclusion}
Multimodal learning is a hot topic in the research community, of which a key challenge is modality missing.
In practice, the multimodal data may not be complete due to various reasons.
Previous works usually cannot effectively utilize the modality-missing data for the learning task.
To address this problem, we propose an efficient approach to leverage the knowledge in the modality-missing data.
Specifically, we present a system based on maximum likelihood estimation to characterize the conditional distribution of the modality-complete data and the modality-missing data, which has a theoretical guarantee.
Furthermore, we develop a generalized form of the softmax function to effectively implement our maximum likelihood estimation framework in an end-to-end way.
We conduct experiments on the eNTERFACE’05 dataset and the RAVDESS dataset for multimodal learning to demonstrate the effectiveness of our approach.
In the future, we will extend our approach to more complex multimodal learning scenarios. 
For example, we can consider that missing modalities exist in both training and testing phases. 
In addition, we can further study the scenario with missing modalities and missing labels.

\bibliographystyle{unsrt}
\bibliography{reference}

\end{document}